\documentclass[11pt,a4paper]{article}

\usepackage{times,latexsym}
\usepackage[hyphens]{url}
\usepackage[T1]{fontenc}
\usepackage{xspace}
\usepackage{amsmath}
\usepackage{multirow}
\usepackage{arydshln}
\usepackage{booktabs}
\usepackage{url}
\usepackage{adjustbox}
\usepackage{amssymb}
\usepackage{subcaption}
\usepackage{dblfloatfix}
\usepackage{relsize}
\usepackage{soul}
\usepackage{siunitx}

\usepackage[acceptedWithA]{tacl2018v2}

\usepackage{xspace,mfirstuc,tabulary}

\newif\iftaclinstructions
\taclinstructionsfalse 
\iftaclinstructions
\renewcommand{\confidential}{}
\renewcommand{\anonsubtext}{(No author info supplied here, for consistency with
TACL-submission anonymization requirements)}
\newcommand{\instr}
\fi

\iftaclpubformat 

\else

\fi

\newcommand{\method}[1]{\textsc{#1}\xspace}
\newcommand{\lstm}{\method{lstm}}
\newcommand{\tdlm}{\method{tdlm}}
\newcommand{\gpt}{\method{gpt{\smaller 2}}}
\newcommand{\bert}{\method{bert}}
\newcommand{\bertc}{\method{bert$_{\text{cs}}$}}
\newcommand{\bertu}{\method{bert$_{\text{ucs}}$}}
\newcommand{\xlnet}{\method{xlnet}}
\newcommand{\xlnetu}{\method{xlnet$_{\text{uni}}$}}
\newcommand{\xlnetb}{\method{xlnet$_{\text{bi}}$}}
\newcommand{\lstmreal}{\method{lstm${^+}$}}
\newcommand{\lstmnone}{\method{lstm${^\varnothing}$}}
\newcommand{\lstmrandom}{\method{lstm${^-}$}}
\newcommand{\tdlmnone}{\method{tdlm${^\varnothing}$}}
\newcommand{\tdlmreal}{\method{tdlm${^+}$}}
\newcommand{\tdlmrandom}{\method{tdlm${^-}$}}
\newcommand{\gptnone}{\method{gpt{\smaller 2}${^\varnothing}$}}
\newcommand{\gptreal}{\method{gpt{\smaller 2}${^+}$}}
\newcommand{\gptrandom}{\method{gpt{\smaller 2}${^-}$}}
\newcommand{\bertcnone}{\method{bert$_{\text{cs}}^\varnothing$}}
\newcommand{\bertcreal}{\method{bert$_{\text{cs}}^+$}}
\newcommand{\bertcrandom}{\method{bert$_{\text{cs}}^-$}}
\newcommand{\bertunone}{\method{bert$_{\text{ucs}}^\varnothing$}}
\newcommand{\bertureal}{\method{bert$_{\text{ucs}}^+$}}
\newcommand{\berturandom}{\method{bert$_{\text{ucs}}^-$}}
\newcommand{\xlnetunone}{\method{xlnet$_{\text{uni}}^\varnothing$}}
\newcommand{\xlnetureal}{\method{xlnet$_{\text{uni}}^+$}}
\newcommand{\xlneturandom}{\method{xlnet$_{\text{uni}}^-$}}
\newcommand{\xlnetbnone}{\method{xlnet$_{\text{bi}}^\varnothing$}}
\newcommand{\xlnetbreal}{\method{xlnet$_{\text{bi}}^+$}}
\newcommand{\xlnetbrandom}{\method{xlnet$_{\text{bi}}^-$}}
\newcommand{\ubone}{\method{ub${_1}$}}
\newcommand{\ubtwo}{\method{ub${_2}$}}
\newcommand{\ubonef}{\method{ub${_1^\varnothing}$}}
\newcommand{\ubtwof}{\method{ub${_2^\varnothing}$}}
\newcommand{\hnone}{\method{h${^\varnothing}$}}
\newcommand{\hreal}{\method{h${^+}$}}
\newcommand{\hrandom}{\method{h${^-}$}}
\newcommand{\modelprob}[1]{P(#1)}
\newcommand{\unigramprob}[1]{P_{\text{u}}(#1)}

\newcommand{\measure}[1]{{\textsl{#1}}\xspace}
\newcommand{\lp}{\measure{LP}}
\newcommand{\ml}{\measure{MeanLP}}
\newcommand{\pl}{\measure{PenLP}}
\newcommand{\nl}{\measure{NormLP}}
\newcommand{\slor}{\measure{SLOR}}

\newcommand{\bb}[2][]{\textbf{#2}}

\newcommand{\secref}[2][]{Section#1~\ref{sec:#2}}

\newcommand{\tabref}[2][]{Table#1~\ref{tab:#2}}
\newcommand{\figref}[2][]{Figure#1~\ref{fig:#2}}

\newcommand{\eqnref}[2][]{Equation#1~(\ref{eqn:#2})}

\usepackage{fancyhdr}
\title{How Furiously Can Colourless Green Ideas Sleep?\\Sentence 
Acceptability in Context}

\author{Jey Han Lau$^{1,7}$ Carlos Armendariz$^{2}$ Shalom
Lappin$^{2,3,4}$  \textbf{Matthew
Purver}$^{2,5}$ \textbf{Chang Shu}$^{6,7}$   \\[1ex]
    $^1$ The University of Melbourne \quad  $^2$ Queen Mary University 
    of
    London \\
    $^3$ University of Gothenburg \quad $^4$ King's College London \\
    $^5$ Jo\v{z}ef Stefan Institute \quad $^6$ University of Nottingham 
    Ningbo
    China \quad
    $^7$ DeepBrain\\[1ex]
    {\footnotesize \sf jeyhan.lau@gmail.com, 
    c.santosarmendariz@qmul.ac.uk} \\
    {\footnotesize \sf shalom.lappin@gu.se, m.purver@qmul.ac.uk, 
 shuchang0011@gmail.com}
 }

\date{}

\begin{document}
\thispagestyle{fancy}
\maketitle
\begin{abstract}
We study the influence of context on sentence acceptability.
First we compare the acceptability ratings of sentences judged in 
isolation, with a relevant context, and with an irrelevant context.  Our 
results show that context induces a cognitive load for
humans, which compresses the distribution of ratings. Moreover, in
relevant contexts we observe a discourse coherence effect which 
uniformly raises acceptability.
Next, we test unidirectional and bidirectional language models in their 
ability to predict acceptability ratings. The bidirectional models show 
very promising results, with the best model achieving a new 
state-of-the-art for unsupervised acceptability prediction.  The two 
sets of experiments provide insights into the cognitive aspects of 
sentence processing and central issues in the computational modelling of 
text and discourse.
\end{abstract}

\section{Introduction}

Sentence \emph{acceptability} is the extent to which a sentence appears natural 
to native speakers of a language. Linguists have often used this property to motivate grammatical 
theories. Computational language processing has traditionally been more concerned with \emph{likelihood} --- the probability of a sentence being produced or encountered. The question of whether and how these properties are related is a fundamental one.
\newcite{Lau+:2017} experiment with unsupervised language models to 
predict acceptability, and they obtained an encouraging correlation with human 
ratings. This
raises foundational questions about the nature of 
linguistic knowledge: if probabilistic models can acquire knowledge of 
sentence acceptability from raw texts, we have prima facie support for 
an alternative view of language acquisition that does not rely on a 
categorical grammaticality component.

It is generally assumed that our perception of sentence acceptability is
influenced by \emph{context}. Sentences which may appear odd in isolation can
become natural in some environments, and sentences which seem perfectly well
formed in some contexts are odd in others. On the computational side, much
recent progress in language modelling has been achieved through the ability to
incorporate more document context, using broader and deeper models
(e.g.\ \newcite{Devlin+:2019,Yang+:2019}). While most language modelling 
is restricted to
individual sentences, models can benefit from using additional context
\cite{khandelwal-etal-2018-sharp}. However, despite the importance of context,
few psycholinguistic or computational studies systematically investigate how
context affects acceptability, or the ability of language models to predict
human acceptability judgments.


Two recent studies which explore the impact of document context on acceptability
judgments both identify a \emph{compression} effect
\cite{Bernardy+:2018,Bizzoni+:2019}. Sentences perceived to be low in
acceptability when judged without context receive a boost in 
acceptability
when judged within context. Conversely those with high out-of-context 
acceptability see a reduction in acceptability when context is 
presented.
It is unclear what causes this compression effect. Is it a result of
cognitive load, imposed by additional processing demands, or is it the consequence of
an attempt to identify a discourse relation between context and sentence?

 
We address these questions in this paper.
To understand the influence of context on human perceptions, we ran 
three crowdsourced experiments to collect acceptability ratings from 
human annotators.
We develop a methodology to ensure comparable ratings for each 
\textit{target sentence} in isolation (without any context), in a
relevant three-sentence context, and in the context of sentences randomly
sampled from another document. Our results replicate the compression 
effect, and careful analyses reveal that both cognitive load and 
discourse coherence are
involved.




%


To understand the relationship between sentence acceptability and 
probability, we conduct experiments with unsupervised language models to 
predict acceptability.  We explore traditional unidirectional 
(left-to-right) recurrent neural network models, and modern 
bidirectional transformer models (e.g.\ \bert).
We found that bidirectional models consistently outperform 
unidirectional models by a wide margin, calling into question the 
suitability of left-to-right bias for sentence processing. Our best 
bidirectional model achieves simulated human performance on the 
prediction task, establishing a new state-of-the-art.

\section{Acceptability in Context}
\label{sec:amt}

\subsection{Data Collection}
\label{sec:data}

To understand how humans interpret acceptability, we require a set of 
sentences with varying degrees of well-formedness. Following previous 
studies \cite{Lau+:2017,Bernardy+:2018}, we use round trip machine 
translation to introduce a wide range of infelicities into naturally occurring 
sentences.

We sample 50 English (target) sentences and their contexts (three 
preceding sentences) from the English Wikipedia.\footnote{We preprocess 
the raw dump with WikiExtractor 
(\url{https://github.com/attardi/wikiextractor}), and collect paragraphs 
that have $\geq 4$  sentences with each sentence having $\geq 5$ words.  
Sentences and words are tokenised with spaCy (\url{https://spacy.io/}) 
to check for these constraints.} We use Moses to translate the target 
sentences into 4 languages (Czech, Spanish, German and French) and then 
back to English.\footnote{We use the pre-trained Moses models from 
\url{http://www.statmt.org/moses/RELEASE-4.0/models/} for translation.} 
This produces 250 sentences in total (5 languages including English) for 
our \textit{test} set.  Note that we only do round trip translation for 
the target sentences; the contexts are not modified.


We use Amazon Mechanical Turk to collect acceptability ratings for the 
target sentences.\footnote{\url{https://www.mturk.com/}.}
We run three experiments where we expose users to different types of 
context. For the experiments, we split the test set into 25 HITs of 
10 sentences. Each HIT contains 2 original English sentences and 8 
round trip translated sentences, which are different from each other and 
not derived from either of the originals. Users are asked to rate the 
sentences for naturalness on a 4-point ordinal scale: bad (1.0), not 
very good (2.0), mostly good (3.0) and good (4.0). We recruit 
20 annotators for each HIT.

In the first experiment we present only the target sentences, without any 
context. In the second experiment, we first show the context paragraph 
(three preceding sentences of the target sentence), and ask users to 
select the most appropriate description of its topic from a list of 
4 candidate topics. Each candidate topic is represented by three words 
produced by a topic model.\footnote{We train a topic model with 
  50 topics on 15K Wikipedia documents with Mallet \cite{McCallum:2002} 
     and infer topics for the context paragraphs based on the trained 
model.}
Note that the context paragraph consists of original English sentences 
which did not undergo translation. Once the users have selected the 
topic, they move to the next screen where they rate the target sentence 
for naturalness.\footnote{Note that we do not ask the users to judge the 
naturalness of the sentence \textit{in context}; the instructions they 
see for the naturalness rating task is the same as the first 
experiment.}
The third experiment has the same format as the second, except that the 
three sentences presented prior to rating are randomly sampled from 
another Wikipedia article.\footnote{Sampled sentences are sequential, 
running sentences.} We
require annotators to perform a topic identification task prior to 
rating the target sentence to ensure that they read the context before 
making acceptability judgements.

For each sentence, we aggregate the ratings from multiple annotators by 
taking the mean. Henceforth we refer to the mean ratings collected from 
the first (no context), second (real context), and third (random context) 
experiments as \hnone, \hreal and \hrandom, respectively. We rolled out 
the experiments on AMT over several weeks and prevented users from 
doing more than one experiment. Therefore a  disjoint group of 
annotators performed each experiment.

\begin{figure*}[t!]
            \begin{subfigure}{0.32\linewidth}
				\centering
                \includegraphics[width=1\textwidth]{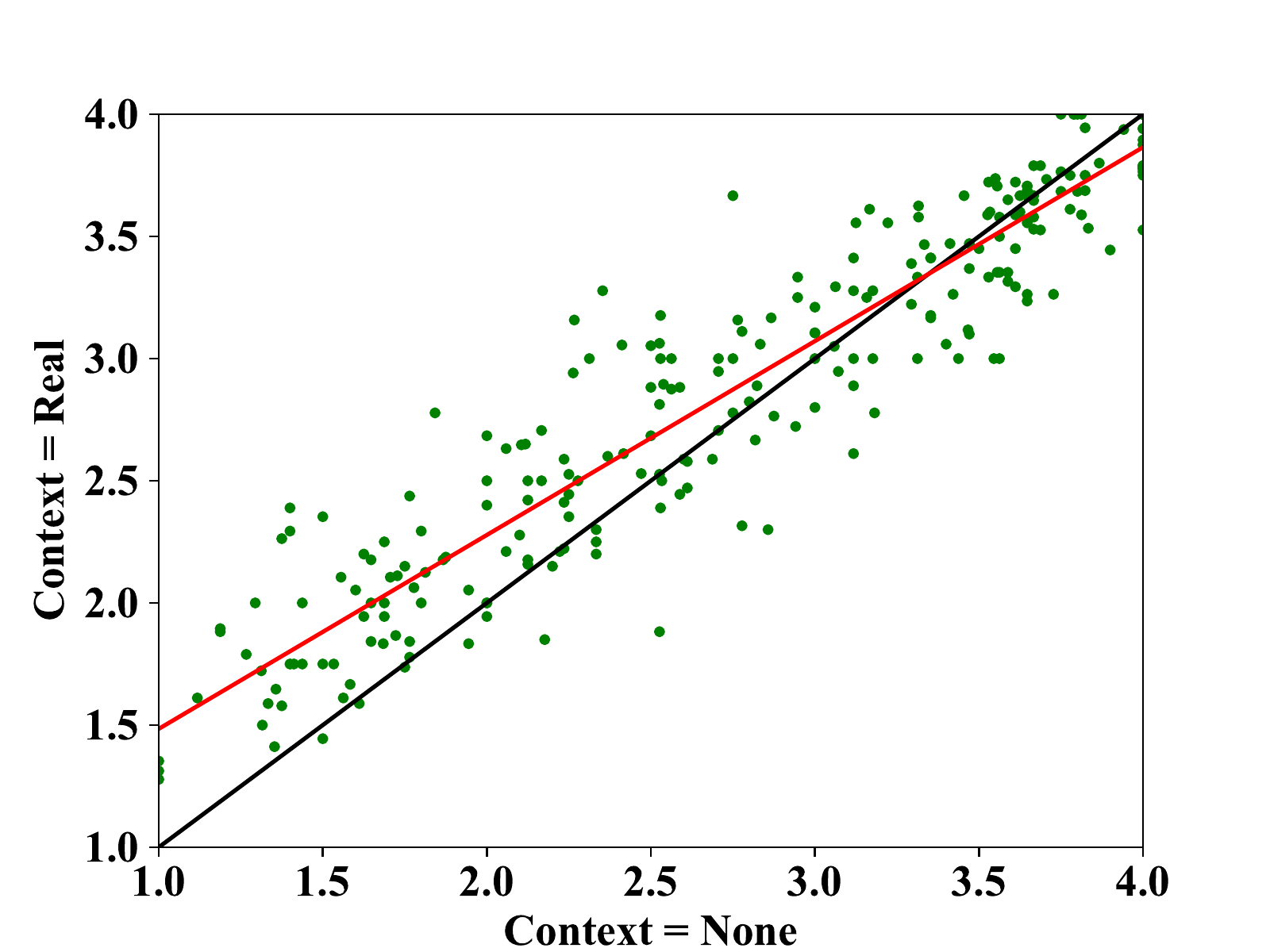}
                \caption{\hreal vs. \hnone}
                \label{fig:sfig1}
			\end{subfigure}
            \begin{subfigure}{0.32\linewidth}
				\centering
                \includegraphics[width=1\textwidth]{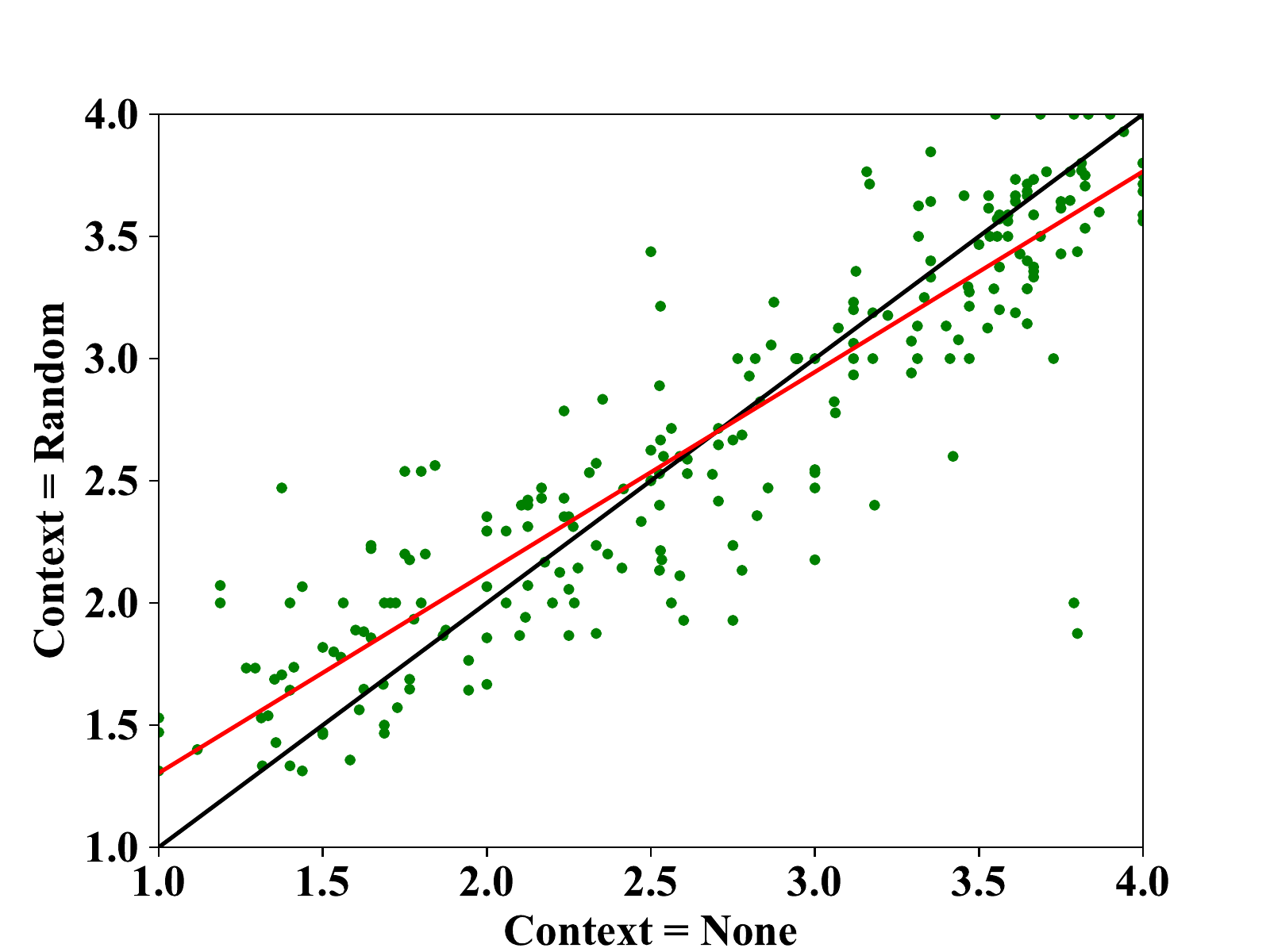}
                \caption{\hrandom vs. \hnone}
				\label{fig:sfig2}
			\end{subfigure}
            \begin{subfigure}{0.32\linewidth}
				\centering
                \includegraphics[width=1\textwidth]{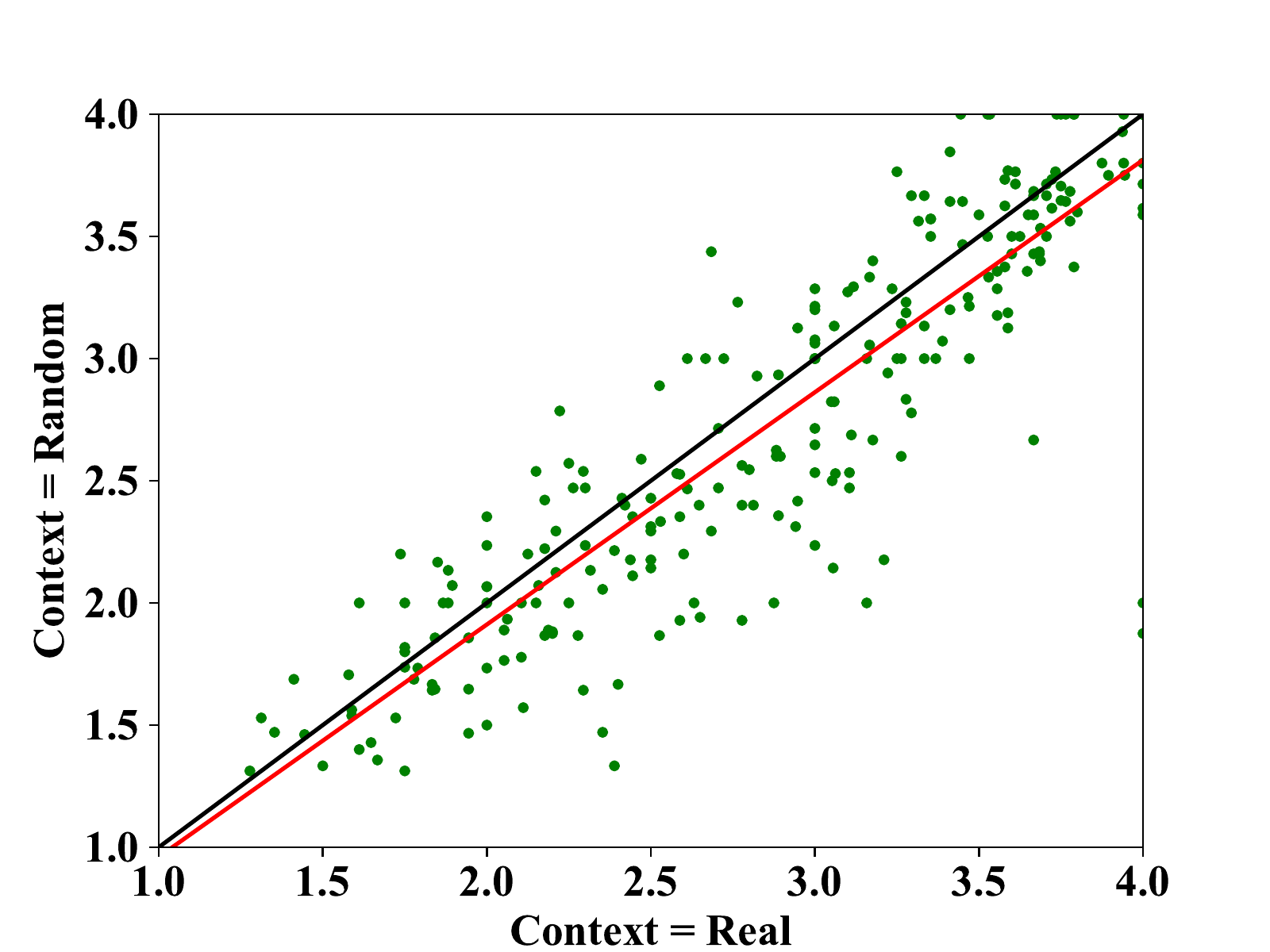}
                \caption{\hrandom vs. \hreal}
				\label{fig:sfig3}
			\end{subfigure}
        \caption{Scatter plots comparing human acceptability ratings.}
		\label{fig:correlation_scatter}
	\end{figure*}

To control for quality, we check that users are rating the English 
sentences $\geq 3.0$ consistently. For the second and third experiments, 
we also check that users are selecting the topics appropriately. In each 
HIT one context paragraph has 1 real topic (from the topic model), and 3 
fake topics with randomly sampled words as the candidate topics.
Users who fail to identify the real topic above a confidence level are 
filtered out.  Across the three experiments, over three quarters of 
workers passed our filtering conditions.

To calibrate for the differences in rating scale between users, we 
follow the postprocessing procedure of \newcite{Hill+:2015}, where we 
calculate the average rating for each user and the overall average (by 
taking the mean of all average ratings), and decrease (increase) the 
ratings of a user by 1.0 if their average rating is greater (smaller) 
than the overall average by 1.0.\footnote{No worker has an average 
rating that is greater or smaller than the overall average by 2.0.} To 
reduce the impact of outliers, for each sentence we also remove ratings 
that are more than 2 standard deviations away from the 
mean.\footnote{This postprocessing procedure discarded a total of 504 
annotations/ratings (approximately 3.9\%) over 3 experiments. The final 
average number of annotations for a sentence in the first, second, and 
third experiments is 16.4, 
17.8, and 
15.3, respectively.}


\subsection{Results and Discussion}

We present scatter plots to compare the mean ratings for the 3 different 
contexts (\hnone, \hreal and \hrandom) in \figref{correlation_scatter}.  
The black line represents the diagonal, and the red one the regression 
line. In general, the mean ratings correlate strongly with each other. 
Pearson's $r$ for \hreal vs.\ \hnone $=$ 0.940, \hrandom vs.\ \hnone $=$ 
0.911, and \hrandom vs.\ \hreal $=$ 0.891.

The regression (red) and diagonal (black) lines in \hreal vs.  
\hnone (\figref{sfig1}) show a compression effect. Bad sentences 
appear a little more natural, and perfectly good sentences become 
slightly less natural when context is introduced.\footnote{On average, 
good sentences (ratings $\geq 3.5$) observe a rating reduction of 0.08 
and bad sentences (ratings $\leq 1.5$) an increase of 0.45.} This is the 
same compression effect observed by \newcite{Bernardy+:2018}. It is also
present in the graph for \hrandom vs.\ \hnone (\figref{sfig2}).

Two explanations of the compression effect seem plausible to us. The 
first is a \textit{discourse coherence} hypothesis that takes this 
effect to be caused by a general tendency to find infelicitous sentences 
more natural in context. This hypothesis, however, does not explain 
why perfectly natural sentences appear less acceptable in 
context. The second hypothesis is a variant of a \textit{cognitive 
load} account. On this view interpreting context imposes a significant burden on a 
subject's processing resources, and this reduces their focus on the 
sentence presented for acceptability judgments. At the extreme ends of 
the rating scale, as they require all subjects to be consistent in order 
to achieve the minimum/maximum mean rating, the increased cognitive load 
increases the likelihood of a subject making a mistake. This
increases/lowers the mean rating, and creates a compression effect.


The discourse coherence hypothesis would imply that the compression 
effect should appear with real contexts, but not with random ones, as 
there is little connection between the target sentence and a random 
context.
By contrast, the cognitive load account predicts that the effect should 
be present in both types of context, as it depends 
only on the processing burden imposed by interpreting the context.   
We see compression in both types of contexts, which suggests that the 
cognitive load hypothesis is the more likely account.

However, these two hypotheses are not mutually exclusive.  It 
is, in principle, possible that both effects --- discourse coherence and 
cognitive load --- are exhibited when context is introduced.


To better understand the impact of discourse coherence, consider
\figref{sfig3}, where we compare \hrandom vs.\ \hreal. Here the 
regression line is parallel to and below the diagonal, implying that 
there is a consistent decrease in acceptability ratings from \hreal to 
\hrandom.  As both ratings are collected with some form of context, the 
cognitive load confound is removed. What remains is a 
discourse coherence effect. Sentences presented in relevant 
contexts undergo a consistent increase in acceptability rating.

To analyse the significance of this effect, we use the non-parametric 
Wilcoxon signed-rank test (one-tailed) to compare the difference between 
\hreal and \ \hrandom. This gives a $p$-value of $1.9 \times 10^{-8}$, 
indicating that the discourse coherence effect is significant.

Returning to Figures \ref{fig:sfig1} 
and \ref{fig:sfig2}, we can see that (1) the offset of 
the regression line, and (2) the intersection point of the diagonal and 
the regression line, is higher in \figref{sfig1}  than in 
\figref{sfig2}. This suggests that there is an increase of ratings, 
and so, in addition to the cognitive load effect, a discourse 
coherence effect is also at work in the real context setting.


%
%
%

We performed hypothesis tests to compare the regression lines in Figures 
\ref{fig:sfig1} and \ref{fig:sfig2} to see if their offsets (constants) 
and slopes (coefficients) are statistically different.\footnote{We 
follow the procedure detailed in 
\url{https://statisticsbyjim.com/regression/comparing-regression-lines/} 
where we collate the data points in Figures \ref{fig:sfig1} and 
\ref{fig:sfig2} and treat the in-context ratings (\hreal and \hrandom) 
as the dependent variable, the out-of-context ratings (\hnone) as the 
first independent variable, and the type of the context (real or random) 
as the second independent variable, to perform regression analyses. The 
significance of the offset and slope can be measured by interpreting the 
$p$-values of the second independent variable, and the interaction 
between the first and second independent variables, respectively.} The 
$p$-value for the offset is $1.7 \times 10^{-2}$, confirming our 
qualitative observation that there is a significant discourse coherence 
effect. The $p$-value for the slope, however, is $3.6 \times 10^{-1}$, 
suggesting that cognitive load compresses the ratings in a consistent 
way for both \hreal and \hrandom, relative to \hnone.


To conclude, our experiments reveal that context induces a cognitive 
load for human processing, and this has the effect of compressing the 
acceptability distribution. It moderates the extremes by making very 
unnatural sentences appear more acceptable, and perfectly natural sentences 
slightly less acceptable. If the context is relevant to the 
target sentence, then we also have a discourse coherence effect, where 
sentences are perceived to be generally more acceptable.

\section{Modelling Acceptability}


In this section, we explore computational models to predict human 
acceptability ratings. We are interested in models that do not rely on 
explicit supervision (i.e.\ we do not want to use the acceptability 
ratings as labels in the training data).  Our motivation here is to 
understand the extent
to which sentence probability, estimated by an unsupervised model, 
can provide the basis for predicting sentence acceptability.

To this end, we train language models (\secref{lm}) using unsupervised 
objectives (e.g.\ next word prediction), and use these models to infer 
the probabilities of our test sentences. To accommodate
sentence length and lexical frequency we experiment with several simple 
normalisation methods, converting probabilities to \textit{acceptability 
measures} (\secref{prob}). The acceptability measures are the final 
output of our models; they are what we use to compare to human 
acceptability ratings.

\subsection{Language Models}
\label{sec:lm}

Our first model is an LSTM language model (\lstm:
\newcite{Hochreiter+:1997,Mikolov+:2010}). Recurrent neural network 
models (RNNs) have been shown to be competitive in this task 
\cite{Lau+:2015,Bernardy+:2018}, and they serve as our baseline.

Our second model is a joint topic and language model (\tdlm: 
\newcite{Lau+:tdlm:2017}). \tdlm combines topic model with language 
model in a single model, drawing on the idea that the topical context of 
a sentence can help word prediction in the language model.  The topic 
model is fashioned as an auto-encoder, where the input is the document's 
word sequence and it is processed by convolutional layers to produce a 
topic vector to predict the input words.  The language model functions 
like a standard LSTM model, but it incorporates the topic vector 
(generated by its document context) into the current hidden state to 
predict the next word.

We train \lstm and \tdlm on 100K uncased English Wikipedia articles 
containing approximately 40M tokens with a vocabulary of 66K 
words.\footnote{We use Stanford CoreNLP \cite{Manning+:2014} to tokenise 
words and sentences. Rare words are replaced by a special UNK symbol.} 

Next we explore transformer-based models, as they have become the benchmark 
for many NLP tasks in recent years 
\cite{Vaswani+:2017,Devlin+:2019,Yang+:2019}. The 
transformer models that we use are trained on a much larger corpus, and they 
are 4--5 times larger with respect to their model parameters.

Our first transformer is \gpt \cite{Radford+:2019}. Given a target 
word, the input is a sequence of previously seen words, which are then 
mapped to embeddings (along with their positions) and fed to multiple 
layers of ``transformer blocks'' before the target word is predicted.  
Much of its power resides in these transformer blocks: each provides a 
multi-headed self-attention unit over
all input words, allowing it to capture multiple dependencies
between words, while avoiding
the need for recurrence. With no need to process a sentence in sequence, 
the model parallelises more efficiently,
and scales in a way that RNNs cannot. 

\gpt is trained on WebText, which consists of over 8 million web 
documents, and uses Byte Pair Encoding (BPE: \newcite{Senrich+:2016}) 
for tokenisation (casing preserved). BPE produces sub-word units, a 
middle ground between word and character, and it provides better coverage 
for unseen words.  We use the released medium-sized model (``Medium'') 
for our experiments.\footnote{\url{https://github.com/openai/gpt-2}.}

Our second transformer is \bert \cite{Devlin+:2019}. Unlike \gpt, 
\bert is not a typical language model, in the sense that it has access 
to both left and right context words when predicting the target 
word.\footnote{Note that \textit{context} is burdened with 2 senses in 
the paper. It can mean the preceding sentences of a target sentence, or 
the neighbouring words of a target word. The intended sense should be 
apparent from the usage.}  Hence, it encodes context in a 
bidirectional manner.

To train \bert, \newcite{Devlin+:2019} propose a masked language model 
objective, where a random proportion of input words are masked and the 
model is tasked to predict them based on non-masked words. In addition 
to this objective, \bert is trained with a next sentence prediction 
objective, where the input is a pair of sentences, and the model's goal 
is to predict whether the latter sentence follows the former. This 
objective is added to provide pre-training for downstream tasks that 
involve understanding the relationship between a pair of sentences, 
e.g.\ machine comprehension and textual entailment.

The bidirectionality of \bert is the core feature that produces its 
state-of-the-art performance on a number of tasks. The flipside of this 
encoding style, however, is that \bert lacks the ability to generate 
left-to-right and compute sentence probability. We discuss how we use 
\bert to produce a probability estimate for sentences in the next 
section (\secref{prob}).

In our experiments, we use the largest pre-trained model 
(``BERT-Large'')\footnote{\url{https://github.com/google-research/bert}.},  
which has a similar number of parameters (340M) to \gpt.  It is trained 
on Wikipedia and BookCorpus \cite{Zhu+:2015}, where the latter is a 
collection of fiction books. Like \gpt, \bert also uses sub-word 
tokenisation (WordPiece). We experiment with two variants of \bert: one 
trained on cased data (\bertc), and another on uncased data (\bertu).  As 
our test sentences are uncased, a comparison between these two models 
allows us to gauge the impact of casing in the training data.

Our last transformer model is \xlnet \cite{Yang+:2019}. \xlnet is 
unique in that it applies a novel permutation language model objective, 
allowing it to capture bidirectional context while preserving key aspects of 
unidirectional language models (e.g.\ left-to-right generation).

The permutation language model objective works by first generating a 
possible permutation (also called ``factorisation order'') of a 
sequence. When predicting a target word in the sequence, the context 
words that the model has access to are determined by the factorisation 
order. To illustrate this, imagine we have the sequence $\mathbf{x} = [ 
x_1, x_2, x_3, x_4 ]$. One possible factorisation order is: $x_3 
\rightarrow x_2 \rightarrow x_4 \rightarrow x_1$. Given this order, if 
predicting target word $x_4$, the model only has access to context words 
$\{x_3, x_2\}$; if the target word is $x_2$, it 
sees only $\{x_3\}$. In practice, the target word is set to be the last 
few words in the factorisation order (e.g.\ $x_4$ and $x_1$), and so the 
model always sees some context words for prediction.

\begin{table*}[t]
\begin{center}
\begin{adjustbox}{max width=\linewidth}
\begin{tabular}{c|c@{\;}c@{\;}c|c@{\;}c@{\;}c@{\;}c}

\toprule
\multirow{2}{*}{\textbf{Model}}
& \multicolumn{3}{c|}{\textbf{Configuration}}
& \multicolumn{4}{c}{\textbf{Training Data}} \\
& \textbf{Architecture} & \textbf{Encoding} & \textbf{\#Param.} & 
\textbf{Casing} & \textbf{Size} & \textbf{Tokenisation} & 
\textbf{Corpora} \\
\midrule

\lstm & RNN & Unidir. & 60M & Uncased & 0.2GB & Word & Wikipedia\\
\tdlm & RNN & Unidir. & 80M & Uncased & 0.2GB & Word & Wikipedia\\
\gpt & Transformer & Unidir. & 340M & Cased & 40GB & BPE & WebText \\
\bertc & Transformer & Bidir. & 340M & Cased & 13GB & WordPiece & 
Wikipedia, BookCorpus\\
\bertu & Transformer & Bidir. & 340M & Uncased & 13GB & WordPiece &  
Wikipedia, BookCorpus\\
\multirow{2}{*}{\xlnet} & \multirow{2}{*}{Transformer} & 
\multirow{2}{*}{Hybrid} & \multirow{2}{*}{340M} & \multirow{2}{*}{Cased} 
& \multirow{2}{*}{126GB} & Sentence- &Wikipedia, BookCorpus, Giga5\\
&&&&&&Piece& ClueWeb, Common Crawl\\
\bottomrule

\end{tabular}
\end{adjustbox}
\end{center}
\caption{Language models and their configurations.}
\label{tab:model-config}
\end{table*}

As \xlnet is trained to work with different factorisation orders during 
training, it has experienced both full/bidirectional context and 
partial/unidirectional context, allowing it to adapt to tasks that have 
access to full context (e.g.\ most language understanding tasks), as well
as those that do not (e.g.\ left-to-right generation).

Another innovation of \xlnet is that it incorporates the segment 
recurrence mechanism of \newcite{Dai+:2019}. This
mechanism is inspired by truncated backpropagation through time used for 
training RNNs, where the initial state of a sequence is initialised with 
the final state from the previous sequence. The segment recurrence 
mechanism works in a similar way, by caching the hidden states of the 
transformer blocks from the previous sequence, and allowing the current 
sequence to attend to them during training. This permits \xlnet to model 
long range dependencies beyond its maximum sequence length.

We use the largest pre-trained model 
(``XLNet-Large''),\footnote{\url{https://github.com/zihangdai/xlnet}.} 
which has a similar number of parameters to our \bert and \gpt models 
(340M). \xlnet is trained on a much larger corpus combining 
Wikipedia, BookCorpus, news and web articles. For tokenisation, 
\xlnet uses SentencePiece \cite{Kudo+:2018}, another sub-word 
tokenisation technique. Like \gpt, \xlnet is trained on cased data.

\tabref{model-config} summarises the language models. In general, the RNN models
are orders of magnitude smaller than the transformers in both model
parameters and training data, although they are trained on the same domain
(Wikipedia), and use uncased data as the test sentences. The RNN models also operate
on a word level, while the transformers use sub-word units.


\subsection{Probability and Acceptability Measure}
\label{sec:prob}

Given a unidirectional language model, we can infer the probability of a 
sentence by multiplying the estimated probabilities of each token using 
previously seen (left) words as context \cite{Bengio+:2003}:
\begin{equation}
\overset{\text{\tiny$\rightarrow$}}{P}(s) = \prod_{i=0}^{|s|} 
P(w_i|w_{<i})
\label{eqn:uni-prob}
\end{equation}
where $s$ is the sentence, and $w_i$ a token in $s$.

\lstm, \tdlm, \gpt are unidirectional models, and so they all compute 
sentence probability as described. \xlnet's unique permutational 
language model objective allows it to compute probability in the same 
way, and to explicitly mark this we denote it as \xlnetu when we infer 
sentence probability using only left context words.

\bert is trained with bidirectional context, and as such it is unable to 
compute left-to-right sentence probability.\footnote{Technically we can 
mask all right context words and predict the target words one at a time, 
but because the model is never trained in this way, we found that 
it performs poorly in preliminary experiments.} We therefore compute 
sentence probability as follows:
\begin{equation}
\overset{\text{\tiny$\leftrightarrow$}}{P} (s) = \prod_{i=0}^{|s|} 
P(w_i|w_{<i}, w_{>i})
\label{eqn:bi-prob}
\end{equation}

With this formulation, we allow \bert to have access to both left and 
right context words when predicting each target word, since this is
consistent with the way in which it was trained. It is important to note, 
however, that sentence probability computed this way is not a \textit{true 
probability value}: these probabilities do not sum 
to 1.0 over all sentences. \eqnref{uni-prob}, in contrast, does guarantee true
probabilities. Intuitively, the sentence probability computed with this 
bidirectional formulation is a measure of the model's confidence in the 
likelihood of the sentence.

To compute the true probability, \newcite{Wang+:2019} show that we 
need to sum the pre-softmax weights for each token to score a sentence, 
and then divide the score by the total score of all sentences. As
it is impractical to compute the total score of all sentences (an infinite set), 
the true sentence probabilities for these bidirectional models are
intractable. We use our non-normalised confidence scores as stand-ins for these
probabilities. 

For \xlnet, we also compute sentence probability this way, applying
bidirectional context, and we denote it as \xlnetb. Note that \xlnetu and 
\xlnetb are based on the same trained model. They differ only in how 
they estimate sentence probability at test time.

\begin{table}[t]
\small
\begin{center}
\begin{tabular}{rl}
\toprule
\textbf{Acc. Measure} & \textbf{Equation} \\
\midrule
\lp & $\log \modelprob{s}$ \\
\ml & $\dfrac{\log \modelprob{s}}{|s|}$ \\
\pl & $\dfrac{\log \modelprob{s}}{((5+|s|)/(5+1))^\alpha}$ \\
\nl & $- \dfrac{\log \modelprob{s}} {\log \unigramprob{s}}$ \\
\slor & $\dfrac{\log \modelprob{s}
- \log \unigramprob{s}}{|s|}$ \\
\bottomrule
\end{tabular}
\end{center}
\caption{{\footnotesize{Acceptability measures for predicting the acceptability of a
sentence; $P(s)$ is the sentence probability, computed using 
\eqnref{uni-prob} or \eqnref{bi-prob} depending on the model;  
$\unigramprob{s}$  is the sentence probability estimated by a unigram 
language model; and $\alpha =$ 
0.8.}
}}
\label{tab:acceptability-measures}
\end{table}

Sentence probability (estimated either using unidirectional or 
bidirectional context) is affected by its length (e.g.\ longer sentences 
have lower probabilities), and word frequency (e.g.\ \textit{the cat is 
big} vs.\ \textit{the yak is big}). To modulate for these factors we 
introduce simple normalisation techniques.  
\tabref{acceptability-measures} presents 5 methods to map sentence 
probabilities to \textit{acceptability measures}: \lp, \ml, \pl, \nl and 
\slor.

\lp is the unnormalised log probability. Both \ml and \pl are normalised 
on sentence length, but \pl scales length with an
exponent ($\alpha$) to dampen the impact of large values 
\cite{Wu+:2016,Vaswani+:2017}. We set $\alpha = 
0.8$ in our experiments. \nl normalises using unigram sentence
probability (i.e.\ $\unigramprob{s} = \prod_{i=0}^{|s|} P(w_i)$), while 
\slor utilises both length and unigram probability \cite{Pauls+:2012}.

When computing sentence probability we have the option of including 
the context paragraph that the human annotators see (\secref{amt}).
We use the superscripts $\varnothing$, $+$, 
$-$ to denote a model using no context, real context, and random 
context respectively (e.g.\ \lstmnone, \lstmreal, and \lstmrandom). Note 
that these variants are created at test time, and are all based on 
the same trained model (e.g.\ \lstm).

For all models except \tdlm, incorporating the context paragraph is 
trivial. We simply prepend it to the target sentence before computing 
the latter's probability. For \tdlmreal or \tdlmrandom, the context 
paragraph is treated as the document context,
from which a topic vector is inferred
and fed to the language model for 
next-word prediction. For \tdlmnone, we set the topic vector to zeros.

\subsection{Implementation}
\label{sec:implementation}

For the transformer models (\gpt, \bert and \xlnet), we use the 
implementation of 
\textit{pytorch-transformers}.\footnote{\url{https://github.com/huggingface/pytorch-transformers}.
Specifically, we employ the following pre-trained models: 
\texttt{gpt2-medium} for \gpt, \texttt{bert-large-cased} for \bertc, 
\texttt{bert-large-uncased} for \bertu, and \texttt{xlnet-large-cased} 
for \xlnetu/\xlnetb.}

\xlnet requires a long dummy context prepended to the target 
sentence for it to compute the sentence probability 
properly.\footnote{In the scenario where we include the context 
paragraph (e.g.\ \xlnetureal), the dummy context is added before it.}  
Other researchers have found a similar problem when using \xlnet  for
generation.\footnote{\url{https://medium.com/@amanrusia/xlnet-speaks-comparison-to-gpt-2-ea1a4e9ba39e}.} 
We think that this is likely due to \xlnet's recurrence mechanism 
(\secref{lm}), where it has access to context from the previous sequence 
during training.

For \tdlm, we use the implementation provided by 
\newcite{Lau+:tdlm:2017},\footnote{\url{https://github.com/jhlau/topically-driven-language-model}.}
following their optimal hyper-parameter configuration without tuning.

We implement \lstm based on Tensorflow's Penn Treebank language 
model.\footnote{\url{https://github.com/tensorflow/models/blob/master/tutorials/rnn/ptb/ptb_word_lm.py}.} 
In terms of hyper-parameters, we follow the configuration of \tdlm where 
applicable. \tdlm uses Adam as the optimiser \cite{Kingma+:2014}, but 
for \lstm we use Adagrad \cite{Duchi+:2011}, as it produces better 
development perplexity.

For \nl and \slor, we need to compute $\unigramprob{s}$, the sentence 
probability based on a unigram language model. As the language models 
are trained on different corpora, we collect unigram counts based on 
their original training corpus. That is,  for \lstm and \tdlm, we use 
the 100K English Wikipedia corpus. For \gpt, we use an open source 
implementation that reproduces the original WebText 
data.\footnote{\url{https://skylion007.github.io/OpenWebTextCorpus/}.} 
For \bert we use the full Wikipedia collection and crawl 
\url{smashwords.com} to reproduce BookCorpus.\footnote{We use the 
scripts in \url{https://github.com/soskek/bookcorpus} to reproduce  
BookCorpus.} Finally, for \xlnet we use the combined set of Wikipedia, 
WebText and BookCorpus.\footnote{\xlnet also uses Giga5 and ClueWeb as 
part of its training data, but we think that our combined collection is 
sufficiently large to be representative of the original training data.}

Source code for our experiments is publicly available at: 
\url{https://github.com/jhlau/acceptability-prediction-in-context}.

\begin{table}[t!]
\begin{center}
\begin{adjustbox}{max width=\linewidth}
\begin{tabular}{c@{\;}c@{\;}c@{\;\;}c@{\;}c@{\;}c@{\;}c@{\;}c}
\toprule
\textbf{Rtg} & \textbf{Encod.} & \textbf{Model} & \textbf{\lp} & 
\textbf{\ml} & \textbf{\pl} & \textbf{\nl} & \textbf{\slor} \\
\midrule

\multirow{16}{*}{\hnone}
&\multirow{8}{*}{Unidir.}
&\lstmnone & 0.29 & 0.42 & 0.42 & 0.52 & \bb{0.53} \\
&&\lstmreal & 0.30 & 0.49 & 0.45 & 0.61 & \bb{0.63} \\
\cdashline{3-8}
&&\tdlmnone & 0.30 &  0.49 &  0.45 &  0.60 &  \bb{0.61} \\
&&\tdlmreal & 0.30 &  0.50 &  0.45 &  0.59 &  \bb{0.60} \\
\cdashline{3-8}
&&\gptnone & 0.33 &  0.34 &  \bb{0.56} &  0.38 &  0.38 \\
&&\gptreal & 0.38 &  0.59 &  0.58 &  \bb{0.63} &  0.60 \\
\cdashline{3-8}
&&\xlnetunone & 0.31 &  0.42 &  0.51 &  0.51 &  \bb{0.52} \\
&&\xlnetureal & 0.36 &  0.56 &  0.55 &  0.61 &  \bb{0.61} \\
\cline{2-8}
&\multirow{6}{*}{Bidir.}
&\bertcnone & 0.51 &  0.54 &  \bb{0.63} &  0.55 & 0.53 \\
&&\bertcreal & 0.53 &  0.63 &  \bb{0.67} &  0.64 & 0.60 \\
\cdashline{3-8}
&&\bertunone & 0.59 &  0.63 &  \bb{0.70} &  0.63 & 0.60 \\
&&\bertureal & 0.60 &  0.68 &  \bb{0.72} &  0.67 &  0.63 \\
\cdashline{3-8}
&&\xlnetbnone & 0.52 &  0.51 &  \bb{0.66} &  0.53 & 0.53 \\
&&\xlnetbreal & 0.57 &  0.65 &  \bb{0.73} &  0.66 & 0.65 \\
\cline{2-8}
&\multirow{2}{*}{---}
&\ubone / \ubonef & &&{0.75 / 0.66} && \\
&&\ubtwo / \ubtwof & &&{0.92 / 0.88} && \\
\midrule

\multirow{16}{*}{\hreal}
&\multirow{8}{*}{Unidir.}
&\lstmnone & 0.29 &  0.44 &  0.43 &  \bb{0.52} & \bb{0.52} \\
&&\lstmreal & 0.31 &  0.51 &  0.46 &  \bb{0.62} & \bb{0.62} \\
\cdashline{3-8}
&&\tdlmnone & 0.30 &  0.50 &  0.45 &  \bb{0.59} & \bb{0.59} \\
&&\tdlmreal & 0.30 &  0.50 &  0.46 &  \bb{0.58} & \bb{0.58} \\
\cdashline{3-8}
&&\gptnone & 0.32 &  0.33 &  \bb{0.56} &  0.36 & 0.37 \\
&&\gptreal & 0.38 &  0.60 &  0.59 &  \bb{0.63} & 0.60 \\
\cdashline{3-8}
&&\xlnetunone & 0.30 &  0.42 &  0.50 &  0.49 & \bb{0.51} \\
&&\xlnetureal & 0.35 &  0.56 &  0.55 &  0.60 & \bb{0.61} \\
\cline{2-8}
&\multirow{6}{*}{Bidir.}
&\bertcnone & 0.49 &  0.53 & \bb{0.62} &  0.54 & 0.51 \\
&&\bertcreal & 0.52 &  0.63 &  \bb{0.66} &  0.63 & 0.58  \\
\cdashline{3-8}
&&\bertunone & 0.58 &  0.63 &  \bb{0.70} &  0.63 & 0.60 \\
&&\bertureal & 0.60 &  0.68 &  \bb{0.73} &  0.67 & 0.63 \\
\cdashline{3-8}
&&\xlnetbnone & 0.51 &  0.50 &  \bb{0.65} &  0.52 & 0.53\\
&&\xlnetbreal & 0.57 &  0.65 &  \bb{0.74} &  0.65 & 0.65 \\
\cline{2-8}
&\multirow{2}{*}{---}
&\ubone / \ubonef & &&{0.73 / 0.66} &&\\
&&\ubtwo / \ubtwof & &&{0.92 / 0.89} &&\\
\midrule

\multirow{16}{*}{\hrandom}
&\multirow{8}{*}{Unidir.}
&\lstmnone & 0.28 &  0.44 &  0.43 &  \bb{0.50} & \bb{0.50} \\
&&\lstmrandom & 0.27 &  0.41 &  0.40 &  \bb{0.47} & \bb{0.47} \\
\cdashline{3-8}
&&\tdlmnone & 0.29 &  0.52 &  0.46 &  \bb{0.59} & 0.58 \\
&&\tdlmrandom & 0.28 &  0.49 &  0.44 &  \bb{0.56} & 0.55 \\
\cdashline{3-8}
&&\gptnone & 0.32 &  0.34 &  \bb{0.55} &  0.35 & 0.35 \\
&&\gptrandom & 0.30 &  0.42 &  \bb{0.51} &  0.44 & 0.41 \\
\cdashline{3-8}
&&\xlnetunone & 0.30 &  0.44 &  \bb{0.51} &  0.49 & 0.49 \\
&&\xlneturandom & 0.29 &  0.40 &  \bb{0.49} &  0.46 & 0.46 \\
\cline{2-8}
&\multirow{6}{*}{Bidir.}
&\bertcnone & 0.48 &  0.53 &  \bb{0.62} &  0.53 & 0.49 \\
&&\bertcrandom & 0.49 &  0.52 &  \bb{0.61} &  0.51 & 0.47 \\
\cdashline{3-8}
&&\bertunone & 0.56 &  0.61 &  \bb{0.68} &  0.60 & 0.56 \\
&&\berturandom & 0.56 &  0.58 &  \bb{0.66} &  0.57 & 0.53 \\
\cdashline{3-8}
&&\xlnetbnone & 0.49 &  0.48 &  \bb{0.62} &  0.49 & 0.48 \\
&&\xlnetbrandom & 0.50 &  0.51 &  \bb{0.64} &  0.51 & 0.50  \\
\cline{2-8}
&\multirow{2}{*}{---}
&\ubone / \ubonef & &&{0.75 / 0.68} &&\\
&&\ubtwo / \ubtwof & &&{0.92 / 0.88} &&\\
\midrule

\end{tabular}
\end{adjustbox}
\end{center}
\caption{Modelling results. Boldface indicates optimal performance in 
each row.}
\label{tab:modelling-results}
\end{table}

\subsection{Results and Discussion}
\label{sec:modelling:results}

We use Pearson's $r$ to assess how well the models' acceptability 
measures predict mean human acceptability ratings, following previous 
studies \cite{Lau+:2017,Bernardy+:2018}. Recall that for each model 
(e.g.\ \lstm), there are 3 variants with which we infer the sentence probability 
at test time. These are distinguished by whether we include no context
(\lstmnone), real context (\lstmreal), or random context (\lstmrandom). There
are also three types of human acceptability ratings (ground truth), 
where sentences are judged with no context, (\hnone), real context (\hreal),
and random context (\hrandom). We present the full results in \tabref{modelling-results}.

To get a sense of what the correlation figures indicate for these 
models, we compute two human performance estimates to serve as upper 
bounds on the
accuracy of a model. The first upper bound (\ubone) is the one-vs-rest annotator 
correlation, where we select a random annotator's rating and compare it to the 
mean rating of the rest, using Pearson's $r$. We repeat this for a large 
number of trials (1000) to get a robust estimate of the mean 
correlation. \ubone can be interpreted as the average human performance 
working in isolation. The second upper bound (\ubtwo) is the 
half-vs-half annotator correlation. For each sentence we randomly split 
the annotators into two groups, and compare the mean rating between 
groups,
again using Pearson's $r$
and repeating it (1000) to get a robust estimate.
\ubtwo can be taken as the average human performance working 
collaboratively. Overall, the simulated human performance is fairly 
consistent over context types (\tabref{modelling-results}), e.g.\ \ubone 
$=$ 
0.75, 
0.73, and 0.75 for \hnone, \hreal, and \hrandom, respectively.

When we postprocess the user ratings, remember that we remove the 
  outlier ratings ($\geq$ 2 standard deviation) for each sentence 
(\secref{data}).  While this produces a cleaner set of annotations, this 
filtering step does (artificially) increase the human agreement or upper 
bound correlations.  For completeness we also present upper bound 
variations where we do not remove the outlier ratings, and denote them 
as \ubonef and \ubtwof. In this setup, the one-vs-rest correlations drop 
to 
0.62--0.66 (\tabref{modelling-results}). Note that all model 
  performances are reported based on the outlier-filtered ratings, 
although there are almost no perceivable changes to the performance 
figures when they are evaluated on the outlier-preserved ground truth.

Looking at \tabref{modelling-results}, the models' performances are 
fairly consistent over different types of ground truths (\hnone, \hreal, 
and \hrandom). This is perhaps not very surprising, as the correlations 
among the human ratings for these context types are very high 
(\secref{amt}).

We now focus on the results with \hnone as ground truth (``Rtg'' $=$ 
\hnone).
\slor is generally the best 
acceptability measure for unidirectional models, with \nl not far behind 
(the only exception is \gptnone).  The recurrent models (\lstm and 
\tdlm) are very strong compared to the much larger transformer models 
(\gpt and \xlnetu). In fact \tdlm has the best performance when context 
is not considered (\tdlmnone, \slor $=$ 
0.61), suggesting that model architecture maybe more important than  
  number of parameters and amount of training data.

For bidirectional models, the unnormalised \lp works very well. The 
clear winner here, however, is \pl. It substantially and consistently  
outperforms all other acceptability measures. The strong performance 
of  \pl that we see here illuminates its popularity in machine 
translation for beam search decoding \cite{Vaswani+:2017}. With the 
exception of \pl, the gain from normalisation for the bidirectional 
models is small, but we don't think this can be attributed to the size 
of models or training corpora, as the  large unidirectional models (\gpt 
and \xlnetu) still benefit from normalisation. The best model without 
considering context is \bertunone with a correlation of 
0.70 (\pl), which is very close to the idealised single-annotator 
  performance \ubone (0.75) and surpasses the unfiltered performance 
\ubonef (0.66), creating a new state-of-the-art for unsupervised 
acceptability prediction \cite{Lau+:2015,Lau+:2017,Bernardy+:2018}.  
There is still room to improve, however, relative to the collaborative 
\ubtwo (0.92) or \ubtwof (0.88) upper bounds.

We next look at the impact of incorporating context at test time 
for the models (e.g.\ \lstmnone vs.\ \lstmreal or \bertunone vs.\ 
\bertureal). To ease interpretability we will focus on \slor for 
unidirectional models, and \pl for bidirectional models. Generally, we 
see that incorporating context always improves correlation, for both 
cases where we use \hnone and \hreal as ground truths, suggesting that 
context is beneficial when it comes to sentence modelling. The only 
exception is \tdlm, where \tdlmnone and \tdlmreal perform very 
similarly. Note, however, that context is only beneficial when it is 
relevant.  Incorporating random contexts (e.g.\ \lstmnone vs.\ 
\lstmrandom or \bertunone vs.\ \berturandom with \hrandom as ground 
truth) reduces the performance for all models.\footnote{There is one 
exception: \xlnetbnone (0.62) vs.\ \xlnetbrandom (0.64). As we saw 
previously in \secref{implementation}, \xlnet requires a long dummy 
context to work, and so this observation is perhaps unsurprising, 
because it appears that context --- whether it is relevant or not --- 
seems to always benefit \xlnet.}

Recall that our test sentences are uncased (an artefact of Moses, the 
machine translation system that we use). While the recurrent models are 
all trained on uncased data, most of the transformer models are trained 
with cased data. \bert is the only transformer that is 
pre-trained on both cased (\bertc) and uncased data (\bertu). To 
understand the impact of casing, we look at the performance of \bertc 
and \bertu with \hnone as ground truth. We see an 
improvement of 5--7 points (depending on whether context is 
incorporated), which suggests that casing has a significant impact on 
performance. Given that \xlnetbreal already outperforms \bertureal (0.73 vs.\ 0.72), even 
though \xlnetbreal is trained with cased data, we conjecture that an 
uncased \xlnet is likely to outperform \bertunone when context is 
not considered.

To summarise, 
our first important result is the
exceptional performance of bidirectional models. It raises the question of
whether left-to-right bias is an appropriate assumption for predicting sentence
acceptability. One could argue that this result may be due to our experimental
setup. Users are presented with the sentence in text, and they have the opportunity
to read it multiple times, thereby creating  an environment that  may 
simulate bidirectional context. We could test this conjecture by 
changing the presentation of
the sentence, displaying it one word at a time (with older words fading 
off), or
playing an audio version (e.g.\ via a text-to-speech system). However, these
changes will likely introduce other confounds (e.g.\ prosody), but we 
believe it is an interesting avenue for future work. 

Our second result is more tentative.  Our experiments seem to indicate 
that model architecture is more important than training or model size.  
We see that \tdlm, which is trained on data orders of magnitude smaller 
and has model parameters 4 times smaller in size (Table 1), outperforms 
the large unidirectional transformer models. To establish this 
conclusion more firmly we will need to rule out the possibility that the 
relatively good performance of \lstm and \tdlm is not due to a cleaner 
(e.g.\ lowercased) or more relevant (e.g.\ Wikipedia) training corpus.  
With that said, we contend that our findings motivate the construction 
of better language models, instead of increasing the number of 
parameters, or the amount of training data.  It would be interesting to 
examine the effect of extending \tdlm with a bidirectional objective.


Our final result is that our best model, \bertu, attains a human-level 
performance and achieves a new state-of-the-art
performance in the task of unsupervised acceptability prediction. Given 
this level of
accuracy, we expect it would be suitable for tasks like assessing 
student essays and the quality of machine translations.

\section{Linguists' Examples}

One may argue that our dataset is potentially biased, as round-trip 
machine translation may introduce particular types of infelicities or 
unusual features to the sentences \cite{Graham+:2019}.  
\newcite{Lau+:2017} addressed this by creating a
dataset where they sample 50 grammatical and 50 ungrammatical sentences 
from \newcite{Adger:2003}'s syntax textbook, and run a crowdsourced 
experiment to collect their user ratings.  \newcite{Lau+:2017} found 
that their unsupervised language models (e.g.\ simple recurrent networks) predict the acceptability of these sentences with similar 
performances, providing evidence that their modelling results are 
robust.

We test our pre-trained models using this linguist-constructed dataset, 
and found similar observations:  \gpt, \bertc and \xlnetb produce a \pl 
correlation of 
0.45, 0.53, and 0.58 respectively. These results indicate that these language models 
 are able to predict the acceptability of these sentences reliably, 
consistent with our modelling results with round-trip translated 
sentences (\secref{modelling:results}). While the correlations are 
generally lower, we want to highlight that these linguists' examples are 
artificially constructed to illustrate specific syntactic phenomena, and 
so this constitutes a particularly strong case of out-of-domain 
prediction. These texts are substantially different in nature from 
the natural text that the pre-trained language models are trained on 
(e.g.\ the linguists' examples are much shorter --- less than 7 words 
on average ---  than the natural texts).


\section{Related Work}

Acceptability is closely related to the concept of grammaticality.  
The latter is a theoretical construction corresponding to syntactic 
well-formedness, and it is typically interpreted as a 
binary  property (i.e.\ a sentence is either grammatical or ungrammatical).  
Acceptability, on the other hand, includes syntactic, semantic, 
pragmatic, and non-linguistic factors, such as sentence length. It is 
gradient, rather than binary, in nature \cite{Denison:2004,Sorace+:2005,Sprouse:2007}.

Linguists and other theorists of language have traditionally assumed that
context affects our perception of both grammaticality \cite{Bolinger68} and
acceptability \cite{Bever70}, but surprisingly little work investigates this
effect systematically, or on a large scale.
Most formal linguists rely heavily on the analysis of sentences taken in isolation. 
However many linguistic frameworks seek to incorporate aspects of context-dependence. 
Dynamic theories of semantics \citep{Heim82, Kamp.Reyle93,Groenendijk.Stokhof90} attempt
to capture intersentential coreference, binding, and scope phenomena. 
Dynamic Syntax \citep{CannEtAl07RoLC} uses incremental tree construction and semantic type
projection to render parsing and interpretation discourse dependent.
Theories of discourse structure characterise sentence coherence in context 
through rhetorical relations \cite{Mann.Thompson88,Asher.Lascarides03}, or by
identifying open questions and common ground \cite{Ginzburg12}. 
While these studies offer valuable insights into a variety of context 
related linguistic phenomena,
much of it takes grammaticality and acceptability to be binary properties. 
Moreover, it is not formulated in a way that permits fine-grained psychological experiments, 
or wide coverage computational modelling. 

Psycholinguistic work can provide more experimentally grounded approaches.
\newcite{Greenbaum76} found that combinations of particular syntactic
constructions in context affect human judgments of acceptability, although the
small scale of the experiments
make it difficult to draw general conclusions.  More recent work 
investigates related effects, but it tends to focus on
very restricted aspects of the phenomenon. For example, 
 \newcite{ZlogarDavidson18} investigate the influence of context on the 
acceptability of gestures with speech, focussing on interaction with 
semantic content and presupposition.  The \emph{priming} literature 
shows that exposure to lexical and
syntactic items leads to higher likelihood of their repetition in production
\cite{ReitterEtAl11}, and to quicker processing in parsing under certain
circumstances \cite{GiavazziEtAl2018}. Frameworks such as ACT-R \cite{ACT-R}
explain these effects through the impact of cognitive activation
on subsequent processing. Most of these studies suggest that
coherent or natural contexts should increase acceptability ratings, given that
the linguistic expressions used in processing become more activated.
\newcite{WarnerGlass1987} show that such syntactic contexts can indeed affect
grammaticality judgments in the expected way for \emph{garden path}
sentences. \newcite{Cowart94} uses comparison between positive and
negative contexts, investigating the effect of contexts containing alternative more or
less acceptable sentences. But he restricts the test cases to specific 
pronoun binding
phenomena. None of the psycholinguistic work investigates acceptability judgments in
real textual contexts, over large numbers of test cases and human subjects.

Some recent computational work explores the relation of acceptability judgments
to sentence probabilities.  \newcite{Lau+:2015,Lau+:2017} show that the 
output of
unsupervised language models can correlate with human acceptability ratings.
\newcite{Warstadt+:2018} treat this as a semi-supervised problem, training a
binary classifier on top of a pre-trained sentence encoder to predict
acceptability ratings with greater accuracy.
\newcite{Bernardy+:2018} explore incorporating context into
such models, eliciting human judgments of sentence acceptability when the
sentences were presented both in isolation and within a document 
context.  They find a compression effect in the distribution of the 
human
acceptability ratings.
\newcite{Bizzoni+:2019} observe a similar effect in a paraphrase 
acceptability task.

One possible explanation for this compression effect is to take it as 
the expression of cognitive load. Psychological research on the cognitive load effect
\citep{journals/cogsci/Sweller88,ito_corley_pickering_2018,10.3389/fnhum.2016.00240,10.1007/978-3-642-42054-2_70}
indicates that performing a secondary task can degrade or distort subjects'
performance on a primary task. This could cause judgments to regress towards
the mean. However, the experiments of \newcite{Bernardy+:2018} and
\newcite{Bizzoni+:2019} do not allow us to distinguish this possibility from a
coherence or priming effect, as only coherent
contexts were considered.
Our experimental setup improves on this by introducing a topic 
identification task and incoherent (random) contexts in order to tease 
the effects apart.

\section{Conclusions and Future Work}

We found that processing context induces a cognitive load for humans, which 
creates a compression effect on the distribution of acceptability ratings. We also 
showed that if the context is relevant to the sentence, a discourse 
coherence effect uniformly boosts sentence acceptability. Our language 
model experiments indicate that bidirectional models achieve better 
results than unidirectional models. The best bidirectional model 
performs at a human level, defining a new state-of-the art for this 
task.

In future work we will explore alternative ways to present sentences 
for acceptability judgments. We plan to extend \tdlm, incorporating
a bidirectional objective, as it shows significant promise. 
It will also be interesting to see if our observations generalise to other 
languages, and to different sorts of contexts, both linguistic and 
non-linguistic.

\section*{Acknowledgements}

We are grateful to three anonymous reviewers for helpful comments
on earlier drafts of this paper. Some of the work described here was 
presented in talks in the seminar of the Centre for Linguistic Theory and 
Studies in Probability (CLASP), University of Gothenburg, December 2019, 
and in the Cambridge University Language Technology Seminar, February 
2020. We thank the participants of both events for useful discussion. 

Shalom Lappin's work on the project was supported by grant 2014-39 from the
Swedish Research Council, which funds CLASP. Santos Armendariz and Purver were
partially supported by the European Union's Horizon 2020 research and innovation
programme under grant agreement No 825153, project EMBEDDIA (Cross-Lingual
Embeddings for Less-Represented Languages in European News Media). The results
of this publication reflect only the authors' views and the Commission is not
responsible for any use that may be made of the information it contains.

\bibliography{tacl2018}
\bibliographystyle{acl_natbib}

\end{document}